\documentclass[letterpaper, 10 pt, conference]{ieeeconf}  
\IEEEoverridecommandlockouts
\usepackage{amsmath,amssymb,amsfonts}
\usepackage{algorithmic}
\usepackage{graphicx}
\usepackage{textcomp}
\usepackage{xcolor}
\usepackage{eqexpl}
\usepackage{autobreak}
\usepackage{comment}

\usepackage{stfloats} 
\usepackage{amsmath}

\usepackage[utf8]{inputenc}
\usepackage{textgreek}
\usepackage{amssymb}
\usepackage{float}
\usepackage{svg}
\usepackage{tikz}
\usepackage{pgfplots}
\usepackage{booktabs}
\usepackage{authblk}
\usepackage{float}
\usepackage{mathtools}

\usepackage[backend=bibtex,style=numeric,natbib=true,style=ieee]{biblatex}
\graphicspath{{Figures/}}
\pgfplotsset{compat=1.17}

\newlength\imagewidth
\newlength\imagescale

\def\BibTeX{{\rm B\kern-.05em{\sc i\kern-.025em b}\kern-.08em
    T\kern-.1667em\lower.7ex\hbox{E}\kern-.125emX}}
    
\usepackage{mdframed}

\usepackage{algorithm}
\usepackage{algorithmic}
\usepackage{amsmath}
\usepackage{subfig}
\usepackage{siunitx}

\begin{document}

\title{Adaptive PD Control using Deep Reinforcement Learning for Local-Remote Teleoperation with Stochastic Time Delays\\
}

\author{Luc McCutcheon$^{1}$ Saber Fallah $^{1}$ 
\thanks{$^{1}$Luc McCutcheon and Saber Fallah are with CAV-Lab, Department of Mechanical Engineering Sciences, University of Surrey,
\thanks{$^{2}$Sotiris Moschoyiannis is with Department of Computer Science, University of Surrey 
        {\tt\small a.tamaddoni-nezhad@surrey.ac.uk}}%
        {\tt\small \{lm01065, s.fallah\}@surrey.ac.uk}}%

}

\maketitle

\begin{abstract}

Local-remote systems allow robots to execute complex tasks in hazardous environments such as space and nuclear power stations. However, establishing accurate positional mapping between local and remote devices can be difficult due to time delays that can compromise system performance and stability. Enhancing the synchronicity and stability of local-remote systems is vital for enabling robots to interact with environments at greater distances and under highly challenging network conditions, including time delays. We introduce an adaptive control method employing reinforcement learning to tackle the time-delayed control problem. By adjusting controller parameters in real-time, this adaptive controller compensates for stochastic delays and improves synchronicity between local and remote robotic manipulators. 

To improve the adaptive PD controller's performance, we devise a model-based reinforcement learning approach that effectively incorporates multi-step delays into the learning framework. Utilizing this proposed technique, the local-remote system's performance is stabilized for stochastic communication time-delays of up to 290ms. Our results demonstrate that the suggested model-based reinforcement learning method surpasses the Soft-Actor Critic and augmented state Soft-Actor Critic techniques. Access the code at:
https://github.com/CAV-Research-Lab/Predictive-Model-Delay-Correction
\end{abstract}

\section{Introduction}
\label{sec:intro}
Remote control of complex systems has become an essential capability in today's interconnected world. Local-remote systems are bilateral teleoperated robotic manipulation devices which provide a means for individuals to interact with environments from remote locations. These systems find widespread use across various industries. For instance, remote surgery enables access to expert surgeons located far away from the patient \cite{choi2018telesurgery}. In nuclear power plants, local-remote systems minimize human exposure to radiation during tasks near the reactor core \cite{wang2008tele}. In space engineering, local-remote systems allow for conducting repairs or experiments on space stations while ensuring the safety of astronauts \cite{ruoff1994teleoperation}.

Local-remote teleoperation provides a framework for an operator to interact with a remote environment using intermediary devices. These intermediary devices consist of two parts: the local device and the remote device. In this work, we focus on local-remote systems consisting of two identical robotic arms linked by position-mapping Proportional-Derivative (PD) controllers. PD controllers are a simplistic closed loop control system common in industrial applications, where a feedback response is generated using coefficients proportional to the error and its derivative.

The local device is controlled by a human operator, while the remote system translates actions into a parallel environment.
One of the key challenges in local-remote teleoperation is the communication time-delay, which can adversely affect control performance. Stochastic variable time delay, in particular, poses a significant challenge to modern control solutions, as existing methods are unable to effectively stabilize the robotic system in highly stochastic and long delays. To address this challenge, we propose using Reinforcement Learning (RL) to stabilize robotic manipulators more effectively and enhance system telepresence.

In a local-remote system, an operator controls a robot arm through a local interface and a remote robot arm mimics the operators actions in a remote location. Action delay and observation delay are the two components of bidirectional time delays that impact the control of a teleoperated local-remote system.  Action delay refers to the time lag between when the operator sends a command and when the robot arm completes the action. For example, if the operator sends a command to move the robot arm forward, action delay would refer to the time it takes for the remote robot to receive and execute the command. On the other hand, observation delay is the duration between when the robot arm captures its current state and when the operator receives the updated information. 

In model-free RL, the agent learns an optimal policy through a series of trial-and-error experiences in an environment without explicitly constructing a model of the environment. A typical approach to enhancing model-free RL performance under time delay is to add a buffer of actions taken over the delay period to the state of the system. However, this approach increases the state space exponentially, making the problem intractable \cite{Nath2021}. Another limitation of model-free methods is that it can be difficult to transfer learnt policies when the delay period changes. This is due to the input dimensions depending on the delay length.

These limitations motivate us to use model-based solutions. Model-based methods attempt to learn the dynamics of the environment explicitly. Existing model-based methods require recursive predictions to obtain a future state estimate. This computational inefficiency can limit the real-world applications \cite{firoiu2018human, otto2013curse, chen2021delay}. Instead, we propose using a computationally efficient predictive model (described in Section \ref{sec:ensemble}) to mitigate the effects of stochastic time delay in control systems, without the use of planning.

Our proposed model-based RL approach provides safe and real-time adaptation to increase the performance of PD controllers in delayed conditions. In this approach, the PD controller parameters are predicted as the output of RL agent \cite{shuprajhaa2022reinforcement} where the agent learns to minimize the error between local and remote devices. By using RL the need for pre-determined manually tuned parameters is alleviated since the system is robust to changes in the environment dynamics \cite{sedighizadeh2008adaptive}. This combination of methods provides some of the safety guarantees of classical control, and the improved performance of RL. 

The work presented in this paper contributes to the literature by providing an RL-based adaptive PD controller which has been optimized specifically for stochastically delayed conditions. 

The main contributions of this paper are:
\begin{itemize}

\item Introduces State-Buffer based State Prediction (SBSP), an efficient framework for predicting future states after time-delay.
\item A model-based RL approach to delayed Markov Decision Processes (MDP), Predictive Model Delay Correction (PMDC), which uses SBSP to address the adverse effects of time delay in control systems.
\item Application to the task of synchronising local-remote systems through the use of an adaptive PD controller.
\item Extension of PMDC to stochastic bi-directional delays by using state augmentation.
\end{itemize}

\section{Related Work}


When dealing with time delays in RL tasks, additional uncertainty arises that can affect the learning process. Typical RL algorithms, such as Soft-Actor-Critic (SAC) \cite{haarnoja2018softa, haarnoja2018softb}, are oblivious to the effects of delayed actions and thus have reduced performance imposed by partial observability. 

A Markov Decision Process (MDP) is a framework for formulating optimisation problems. It necessitates that a state is Markovian, implying that an agent is equipped with all requisite information for decision-making at any specific state. In order to make the MDP fully observable, an augmented approach \cite{Nath2021} has been proposed in which past actions taken during the delay period are incorporated into the state information. This approach enables the derivation of an optimal policy \cite{bouteiller2020reinforcement} by ensuring the Markov property, but it comes at the cost of the state space expanding as the delay length increases.

Additionally, it is important to note that the actions that are augmented to the state are off-policy and do not represent a trajectory chosen under the current policy. This can lead to suboptimal results in practice. To address this issue, Bouteiller et al. \cite{bouteiller2020reinforcement} proposed a method that resamples off-policy action trajectories into on-policy trajectories. However, this trajectory resampling requires additional computation as the delay increases.

To overcome these limitations, researchers have explored alternative representations of the action buffer. For instance, Liotet et al. \cite{Liotet2021} introduced a belief representation of the action history. This technique appends a condensed belief representation of the action buffer to the state information instead of the entire action buffer. This belief representation limits the augmented state representation dimensiality and thus the problem complexity. However this representation offers only an approximation of the action-history which decreases observability over complete information solutions.

Due to the limitations of model-free methods, model-based approaches were developed to predict the state in a delayed environment, where a model estimates the transition dynamics \cite{walsh2009learning, chen2021delay}. To do this a model can recursively \emph{undelay} the action delays \cite{firoiu2018human, chen2021delay}, however, this has poor computational complexity due to the numerous model predictions used to calculate future states. 

In addition, the majority of literature focuses solely on constant delays \cite{altman1992closed,bander1999markov,brooks1972markov,kim1985state,kim1987partially}, with planning methods unable to directly address stochasticity in random delays. The work relating to stochastic delays are model-free approaches which use state augmentation \cite{bouteiller2020reinforcement, Nath2021, katsikopoulos2003markov}.
It must be noted the stochastically delayed MDP used in our research differs from \cite{katsikopoulos2003markov} to avoid unrealistic assumptions about delayed conditions, which limit application.

To align with the current architecture and provide additional safety in delayed conditions we apply Predictive Model Delay Correction (PMDC) to the task of real-time adaptive PD controller tuning. Adaptive control of Proportional-Integral-Derivative (PID) based controllers using RL has been explored in previous work \cite{shuprajhaa2022reinforcement, lee2020reinforcement,sun2022reinforcement,wang2007proposal} where RL predicts the PID based controller parameters at each time-step. This approach was first proposed by Wang et al. in 2007 \cite{wang2007proposal}, who demonstrated its effectiveness in controlling a chaotic system. 
Our work contributes to the existing literature by presenting an adaptive PD controller that overcomes time delays via the application of PMDC, thereby filling a gap in the literature. Although a PD controller is used in our application, the proposed methodology can be generalized to other feedback control schemes, such as PID or similar variations, with similar outcomes.

\section{Prerequisites}

 


\subsection{Markov Decision Process}
\label{sec:MDP}
An MDP is a mathematical framework for modelling decision-making systems where the next state is only dependent on the current state and action. It can be defined as a tuple $MDP = (S, A, \mu, p)$ where:

\begin{itemize}
 \item $S$ is the state space, which is the set of all possible states of the system.
 \item $A$ is the action space, which is the set of all possible actions that the agent can take.
 \item $d_0(s_0): S \rightarrow \mathbb{R}$ is the initial state distribution, which is a probability distribution over the states that describes the starting state of the system.
 \item $p(s', r | s, a): S \times A \rightarrow \mathbb{R}$ is the transition probability function, which describes the probability of transitioning from state $s \in S$ to state $s' \in S$ and receiving reward $r$ when taking action $a \in A$.
\end{itemize}
This paper assumes knowledge of the reward function $r$, but does not assume knowledge of the transition probability function $p$. When action delay, denoted by $\alpha$, is applied to the MDP an agent will enact a trajectory between action selection and execution (\ref{eq:trajdista}). This increases the complexity of the task since the state of the system may have changed considerably under the effect of previous actions in the trajectory before the effective action is executed. In this paper a subscript or superscript $R$ is used to distinguish variables relating to the remote system and $L$ for variables relating to the local system when appropriate, when no distinction is made it is assumed to be relating to the adaptive-control agent used on the remote system.

From this definition, we can derive a distribution of trajectories, denoted by $\tau$, that an agent may follow under $\alpha$, before the action is actually applied.

\begin{equation}
    \label{eq:trajdista}
    \tau_{\pi} = d_0(s_0) \prod_{t=0}^{\alpha} \pi(a_t|s_t)p(s_{t+1}|s_t,a_t)
\end{equation}
Where $\pi$ represents the agent's policy

The distribution of trajectories that an agent may follow during the observation delay, denoted by $\omega$, before the observation is returned to the agent, is given by:

\begin{equation}
    \label{eq:trajdistw}
    \tau_{\pi} = d_{\alpha}(s_{\alpha}) \prod_{t=\alpha}^{\alpha + \omega} \pi(a_t|s_t)p(s_{t+1}|s_t,a_t)
\end{equation}

where $d_{\alpha}(s_\alpha)$ represents the state distribution after $\alpha$.
 Agents struggle to learn how one state transitions into another since increasing delay causes agents to enact longer trajectories between selecting an action and receiving a reward. Additionally, as the duration of the delay increases, the agent is confronted with the credit assignment problem \cite{minsky1961steps}, whereby it becomes progressively challenging to attribute credit to individual actions as other actions are performed during the delay period, rendering it difficult to discern which actions correspond to which rewards.

\subsection{Augmented Markov Decision Process}
\label{sec:aug}
The regular MDP formulation, when applied to delayed problems can lead to arbitrarily suboptimal policies \cite{singh1994learning} due to partial observability. In order to ensure the Markov property the Augmented MDP is proposed:

\begin{itemize}
 \item $\mathcal{X} = S \times A^n$ is the state space, where $n$ is the total delay at a given time step.
 \item $A$ is the action space, which is the set of all possible actions that the agent can take.
\item $\delta_0(s_0) = \delta(s_0,a_0,...,a_{n-1}) = \delta(s_0)\prod^{n-1}_{i=0}\delta(a_i-c_i)$\\ is the initial state distribution, where $\delta$ is the Dirac delta function. If $y \sim \delta(\cdot - x)$ then $y = x$ with probability one. $(c_i){i=1}^{n-1}$ denotes the initial sequence of actions.
 \item $p(s', r | s, a): S \times A \rightarrow \mathbb{R}$ is the transition probability function, which describes the probability of transitioning from state $s \in S$ to state $s' \in S$ and receiving reward $r$ when taking action $a \in A$.
\end{itemize}

The state information of the augmented MDP contains the sequences of actions which have not received corresponding observations due to action and observation delay.

\subsection{Time-delayed RL}
Delayed environments use an augmented state space as described in Section \ref{sec:aug} with delayed dynamics. Traditional algorithms such as SAC will always work in randomly delayed conditions. However, their performance will deteriorate because of the added difficulty in credit assignment caused by delayed observations and rewards, along with the exploration and generalization burdens of delayed environments.



\section{Local-Remote Control}
This section outlines a technique for converting a standard RL environment into a local-remote system equivalent. In a local-remote system, the local operator executes the task, while a remote agent runs concurrently and aims to track the operator's path despite the presence of stochastic time delays.

\begin{figure}[!htb]
    \centering
    \includegraphics[scale=0.26]{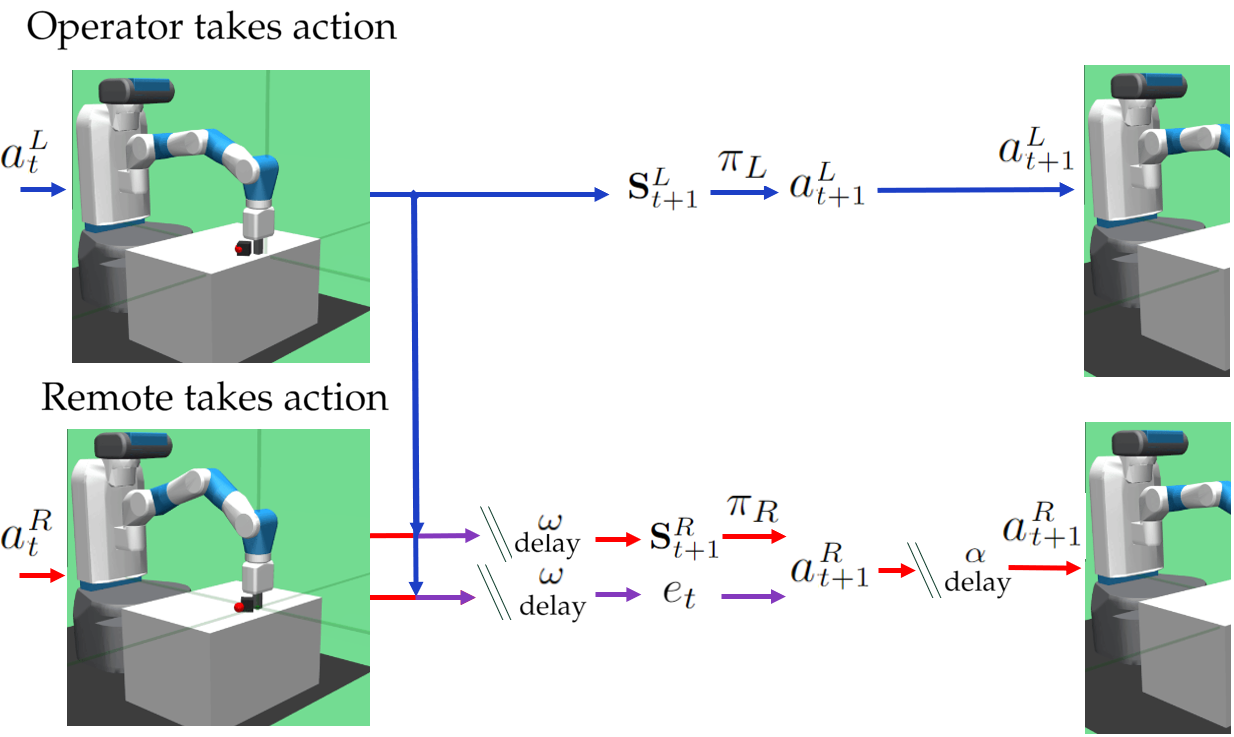}
    \caption{Local-Remote environment architecture, the purple arrows indicate the information from the local device mixing with the remote device (\ref{eq:matrix}). The local policy $\pi_L$ represents the human operator policy, or for ease of training, a pre-trained expert RL policy. $\pi_R$ represents the remote agents policy, where the error (\ref{eq:reward}) between devices not only provides an reward signal, but is also used in (\ref{eq:PID}) to generate action values}
    \label{fig:lr-arch}
\end{figure}

The process begins with the operator executing an action in the local environment. Since the operator is acting in the local environment it has access to all state information concerning itself and the target object.

The observation is first transmitted from the operator to the remote environment, after which an observation delay is applied. Subsequently, the remote observation is created by replacing the remote agent's target information with the operator's location, as described by (\ref{eq:matrix}). 

\begin{equation}
\label{eq:matrix}
s_L \gets \begin{bmatrix}
x_L & y_L & z_L \\
\Dot{x_L} & \Dot{y_L} & \Dot{z_L} \\
x_T & y_T & z_T
\end{bmatrix},
\hspace{2em}
s_R \gets \begin{bmatrix}
x_R & y_R & z_R \\
\Dot{x_R} & \Dot{y_R} & \Dot{z_R} \\
x_L & y_L & z_L
\end{bmatrix}.
\end{equation}

Where $S_L$ and $S_R$ represent the state information of each agent. $x_R$, $y_R$ and $z_R$ represent the remote agent's end-effector position, while $x_L$, $y_L$ and $z_L$ represent the corresponding values for the local device end-effector, and $x_T$, $y_T$ and $z_T$ represent the target's position.

After the remote observation is calculated it is then fed into the SAC controller which outputs PD controller parameters $K_p$ and $K_d$. The error of the system (\ref{eq:reward}) is calculated through the euclidean distance between the operator and remote end-effector positions. This error $e_t$ is used as a reward signal and in the PD controller (\ref{eq:PID}). 

\begin{equation}
\label{eq:reward}
 e_t = - \sqrt{(x_R - x_L)^2+(y_R-y_L)^2+(z_R-z_L)^2}
\end{equation}

\begin{equation}
\label{eq:PID}
    a_t = K_P (e_t) + K_D  \frac{\partial (e_t)}{\partial t},
\end{equation}
 The PD controller takes in $K_P$ and $K_D$ from SAC and the error from (\ref{eq:reward}). The output of the PD controller $a_t$ is the action vector to be applied to the environment and represents forces to be applied to the remote end effector.

\section{Method}
In this section, we present the implementation of PMDC. We first provide an overview of the challenges PMDC aims to solve, we then compare against prior work and provide further details on implementation.



\label{sec:dynamics}
Delayed environments require knowledge of previous actions in order to be fully observable. Appending the action history over the delay period is one method frequently used in model-free literature but rapidly deteriorates in performance as time delay increases. Model-based methods offer better performance by recursively predicting the effect of actions in the action buffer and providing the future state to the RL algorithm. There are two downsides of current model-based approaches to time delay, which PMDC aims to address: 1) recursive model calls can lead to long computation times. 2) the model is only able to handle constant delays.

PMDC addresses problem 1 by introducing State-Buffer based State Prediction (SBSP) as it reduces the number of model predictions per episode. SBSP solves the problem by storing previously calculated future states that subsequent time steps can use. Initially, at training time step 0, PMDC predicts $\alpha$ steps into the future and stores each prediction in a state-buffer. At training time step 1, PMDC uses this stored prediction to make only one additional prediction to see $\alpha+1$ steps into the future from the starting state, which corresponds to the second step after the action delay. This process continues until the end of the episode, with PMDC updating the future state each time step, based on the action chosen by RL. This approach allows for $\alpha$ predictions at time step 0, then one prediction per subsequent time step until the end of the episode. In contrast, the prior method for delay corrected state prediction, hereby referred to as Action-Buffer based State Prediction (ABSP), calculates $\alpha$ predictions per time step. ABSP has been utilized in a number of previous studies \cite{walsh2009learning,derman2021acting,chen2021delay}. Notably, Firoiu et al. had to restrict delay in their experiments due to the escalating computational complexity associated with ABSP \cite{firoiu2018human}.

SBSP bootstraps predictions from the initial state which can lead to accumulating errors towards the end of the episode. To correct these errors, SBSP stores all predicted future states. When it receives the true state of the system observed after the delay period, SBSP calculates the error between the predicted future state $\hat{s}_t$ and the true state $s_t^R$ ($\hat{s}_t^R - s_t^R$). SBSP then deducts this error from its list of calculated future states, including the current prediction for the future state of the system $\alpha$ steps ahead.



\begin{figure}
    \centering
\includegraphics[scale=0.285, trim=0 0 0 0.08cm, clip]{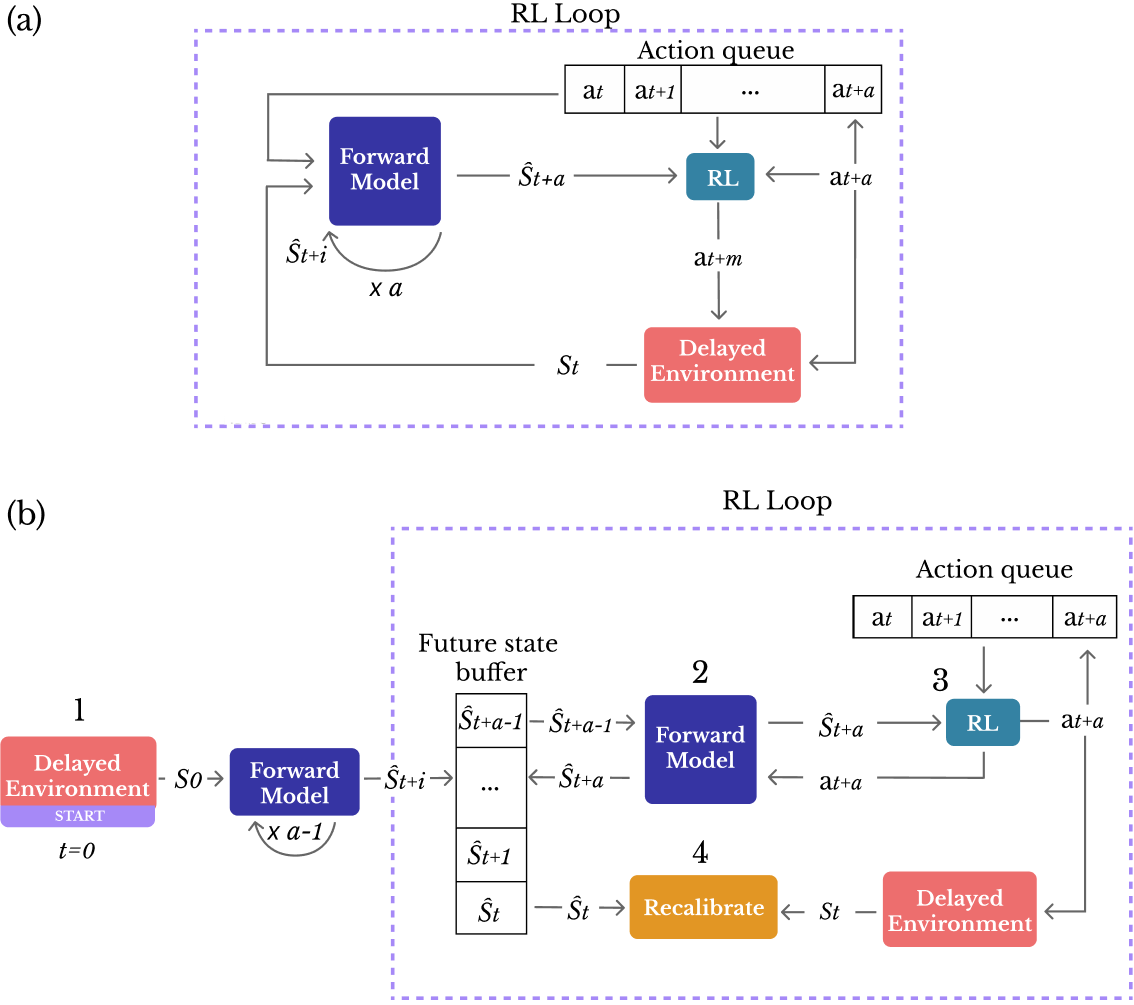}
    \caption{(a) ABSP training loop (b) SBSP training loop}
    \label{fig:new}
\end{figure}

Fig. \ref{fig:new} shows a visual comparison of the (a) ABSP  for providing a delay-corrected state to the RL agent with (b) SBSP our proposed method. The RL loop refers to the iterative process of updating the RL behavior policy, which occurs $T$ times per episode, where $T$ represents the total number of time steps in the episode. The prior method, ABSP, uses a buffer of actions to calculate a corresponding future state each time-step which requires extensive computation where one episode requires $\alpha \times T$ model predictions.

In contrast, the proposed method reuses previous predictions when calculating the future state requiring only $\alpha + T$ for model predictions. SBSP can be interpreted as 4 stages (labelled in Fig. \ref{fig:new} accordingly): 1) Initial delay correction at the beginning of an episode to estimate state $\hat{s}_{\alpha+t-1}$ using ensemble predictive model;  2) Starts the loop used for the remaining time steps by determining the next future state $\hat{s}_{\alpha+t}$ using the predictive model; 3)  Applies the chosen action from RL to the delayed environment; 4) Recalibrates the future state buffer by calculating the error between the predicted and observed states and subtracting this error from the stored states in the future state buffer. Then the the predictive model parameters are updated using the loss between predicted and observed states.

In order to addresses problem 2, stochastic delays, PMDC uses state augmentation. However, in PMDC, only actions taken within the stochastic range are incorporated into the state information. This differs from the state augmentation used in Augmented State SAC (A-SAC), where actions are added over the entire delay length. By only using state augmentation over the stochastic range, the amount of actions added to the state space is greatly reduced compared to A-SAC.

\begin{algorithm}[!htb]
\caption{State-Buffer based State Prediction}
\begin{algorithmic} 
\label{alg:PMDC}
\STATE Initialize replay memory $\mathcal{D}$ to capacity $\mathcal{N}$
\STATE Initialize future state buffer $\mathcal{F}$ to capacity $\alpha$
\STATE Initialize number of models $m$
\STATE Initialize action delay $\alpha$
\STATE Number of episodes $E$

\FOR{episode $= 1,E$}
    \STATE $\mathcal{F}_0 = s_0$
    \FOR{$i=1$ to $\alpha$}
    \STATE \vspace{3pt}$\mathcal{F}_{i} \leftarrow \frac{\sum_{n=0}^{n=m} M_{n}(\mathcal{F}_{i})}{m}$
    \ENDFOR
    \FOR{$t=0, \textit{T}$}

        
        \STATE $a_t \leftarrow \pi(\mathcal{F}_{t+\alpha})$ 
        \STATE Apply action $a_t$ to environment and observe $s_{t+1}$
        \STATE Store \{$\mathcal{F}_t, a_t, \mathcal{F}_{t+1}$\} in $\mathcal{D}$ if transition is non-terminal

        \STATE $\Delta s \leftarrow s_{t+1} - \mathcal{F}_{t+1}$
        \FOR{$i=t+1,t+\alpha$}
            \STATE \vspace{3pt} $\mathcal{F}_i \leftarrow \mathcal{F}_i + \Delta s$
        \ENDFOR
    \STATE $\mathcal{F}_{t+\alpha+1} \leftarrow \frac{\sum_{n=0}^{n=m} M_{n}(\mathcal{F}_{t+\alpha})}{m}$
    \ENDFOR
    \STATE Sample random indexes $i$ where $0<i<\mathcal{N}$
    \FOR{$n=1$ to $m$}
        \STATE Calculate $\mathcal{L}(\mathcal{F}_{i+\alpha},s_i)$ and perform Gradient Decent on $M_n$
    \ENDFOR
\ENDFOR

\end{algorithmic}
\end{algorithm}
Algorithm \ref{alg:PMDC} outlines the steps involved in the SBSP algorithm, this algorithm uses an augmenting state for $s_t$ when used for stochastic delays. In the stochastic case, the state $s_t$ contains the action history over the entire stochastic range. For example, if the delay is 8-12 time steps, the state would include the 4 previous actions. In this algorithm, $\pi$ represents the behavioral policy learned through RL, $s_t$ represents the state at time step $t$, and $M_i$ denotes each model in the ensemble, with $i$ representing its index in the collection.  $\mathcal{F} = ({\hat{s}_t,\hat{s}_{t+1},...,\hat{s}_{t+\alpha}})$ is a buffer that stores previously calculated future state predictions $\hat{s}_t$ from the current time step $t$ to $\alpha$, where $\hat{s}_{t} \approx s_{t}$.



\subsection{Ensemble Learning}
\label{sec:ensemble}
Ensemble learning is a technique that averages the predictions of multiple models to improve accuracy. The benefit of using an ensemble \cite{sagi2018ensemble} is that varying initial weights allow for slight differences in model predictions with their average being a more robust and accurate prediction than any individual model alone.

This work employs 5 feed-forward neural network models with varying initial weights for the ensemble, as this number has been found to enhance performance while maintaining reasonable computational requirements. While it offers the possibility of enhancing performance, the trade-off is increased computational complexity that grows linearly with the number of models used.
    
The predictive model in this work is an ensemble used to predict the future state's after the delay.
\begin{equation}
    \label{eq:huber}
    \mathcal{L}(s_i,\hat{s}_{i+1}) =
        \begin{cases}
            \frac{1}{2}(s_i-\hat{s}_i)^2, & if |s_i-\hat{s}_i|<\delta \\
            \delta (|s_i-\hat{s}_i|-\frac{1}{2}\delta, & otherwise.\\
            
        \end{cases}
\end{equation}
Where $\delta$ represents the threshold for switching between L1 and L2 loss functions, $s_i$ is the observed state, and $\hat{s}_i$ is the model-predicted state.

\begin{figure}
    \centering
    \includegraphics[scale=0.5]{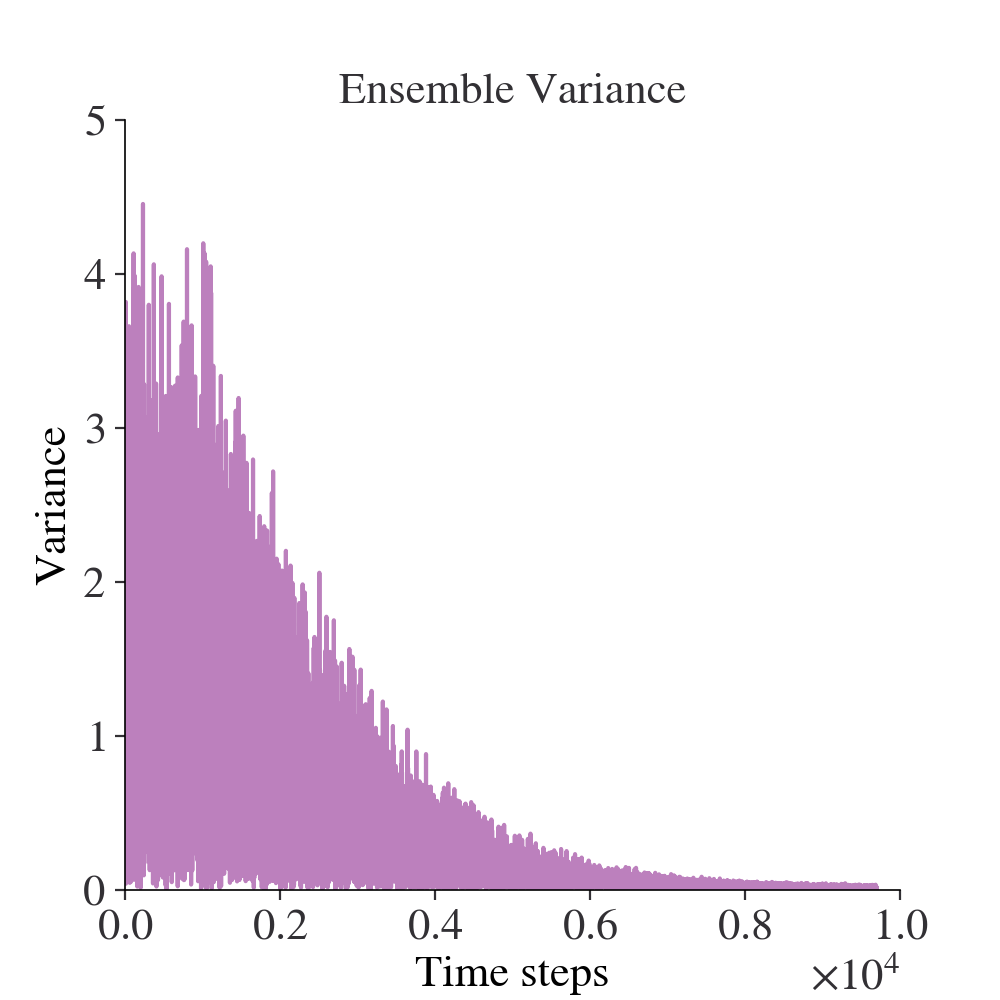}
    \caption{Ensemble Learning variance over initial training period}
    \label{fig:variance}
\end{figure}

This work uses neural networks to approximate the transition function because of their ability to handle the non-linearities present in system dynamics. We use Huber Loss (\ref{eq:huber}) \cite{huber1964robust} to calculate the prediction error before backpropagation as it provides stable robust convergence. The variance over time shows (Fig. \ref{fig:variance}) how the neural networks learn from their predictions and consolidate estimations.


\begin{figure*}[!htb]
\centering

\begin{tabular}{ccc}\\
\includegraphics[trim=15mm 0mm 10mm 5mm, scale=0.39]{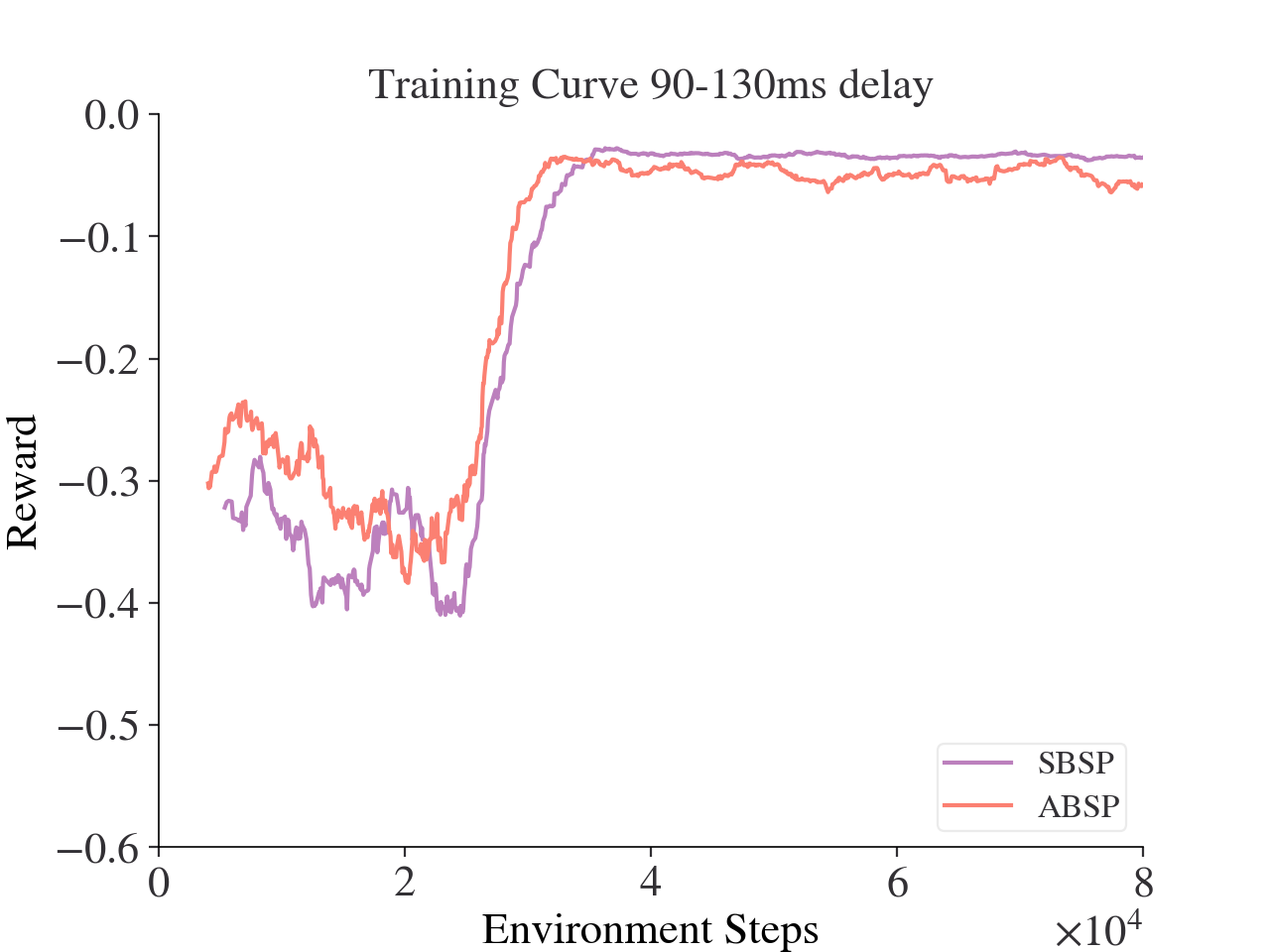} &
\includegraphics[trim=15mm 0mm 10mm 5mm, scale=0.39]{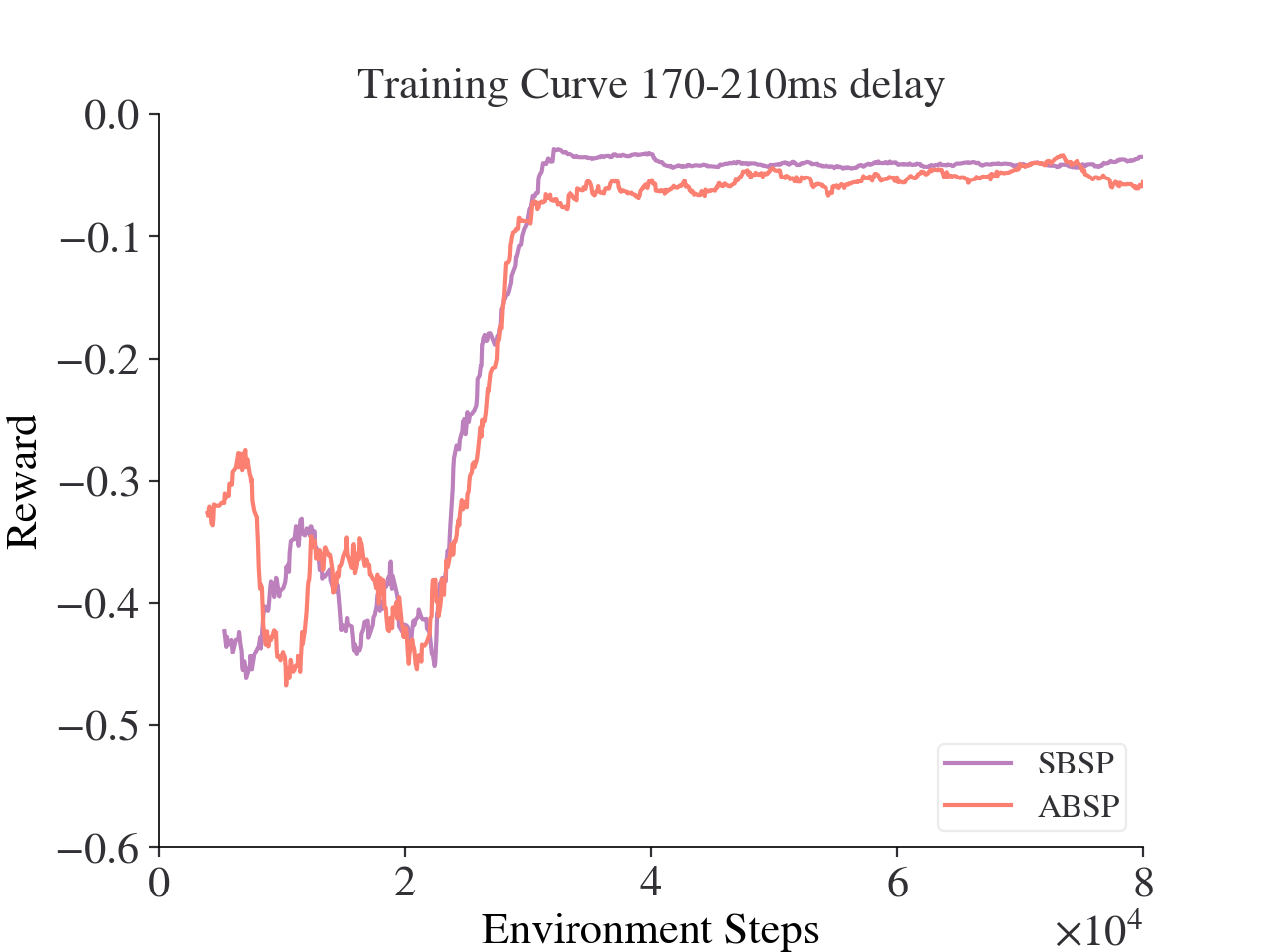} &
\includegraphics[trim=15mm 0mm 12mm 5mm, scale=0.39]{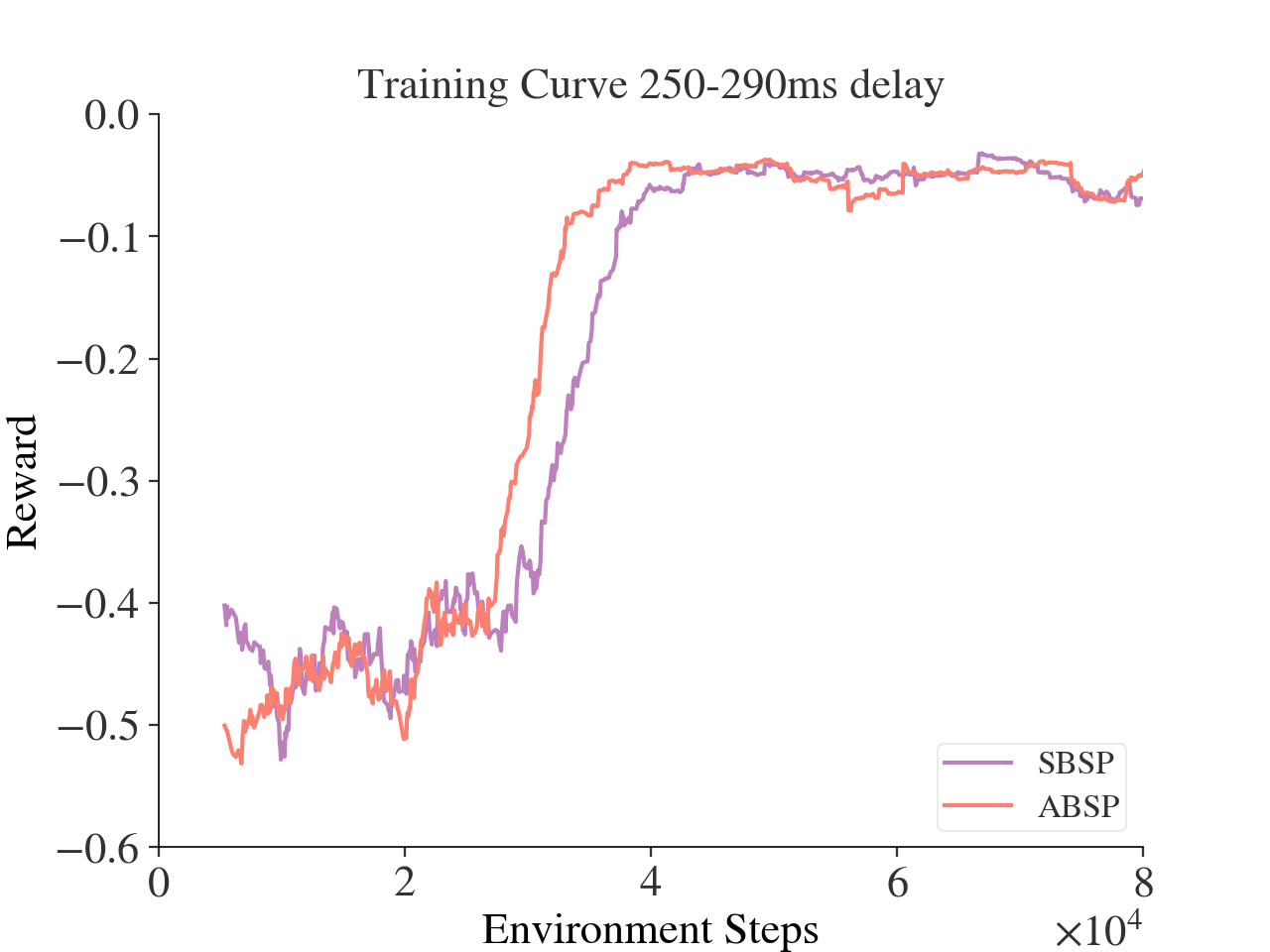} \\
(a) & (b) & (c)\\
\\

\includegraphics[trim=15mm 0mm 10mm 5mm, scale=0.39]{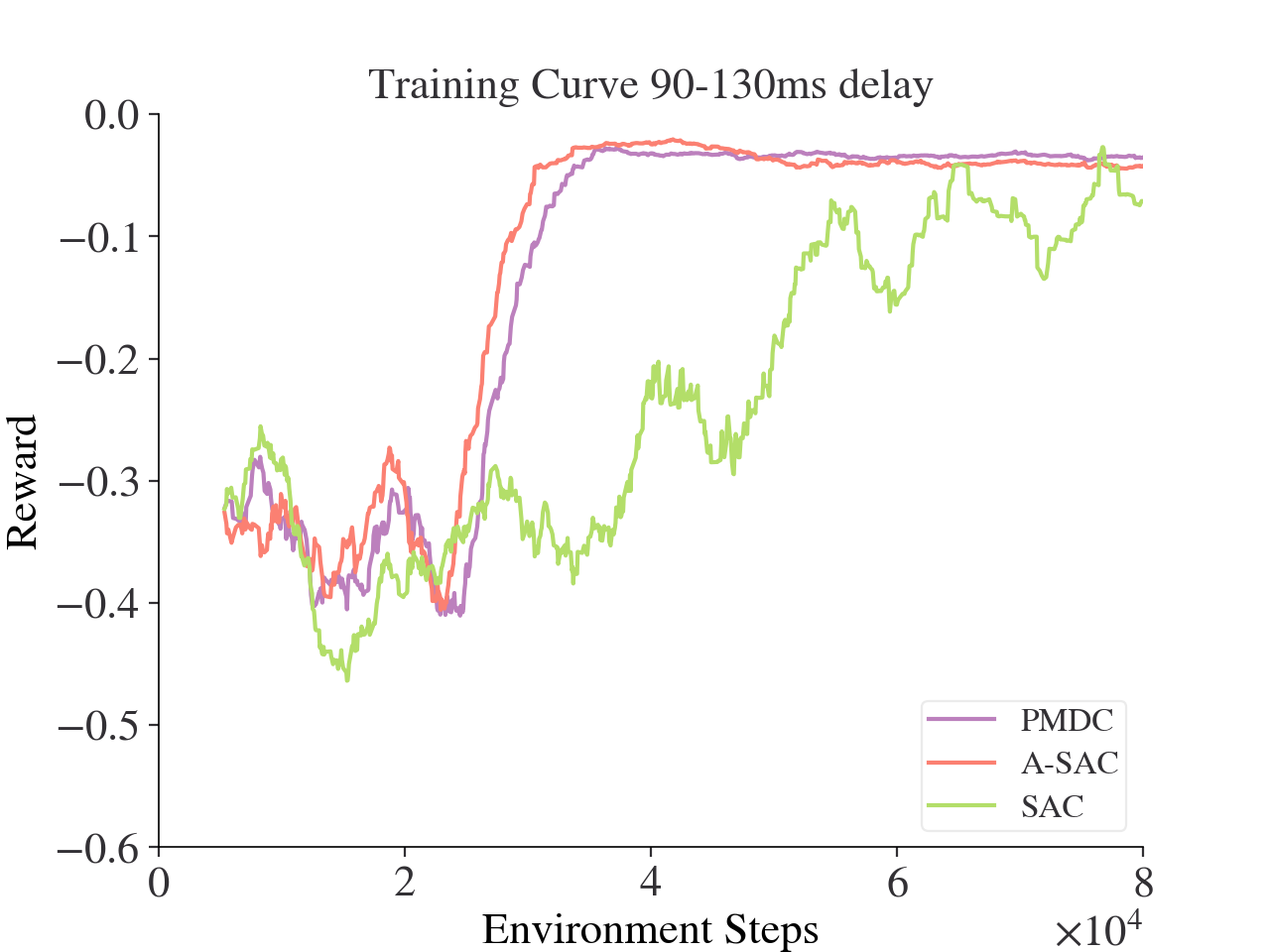} &
\includegraphics[trim=15mm 0mm 10mm 5mm, scale=0.39]{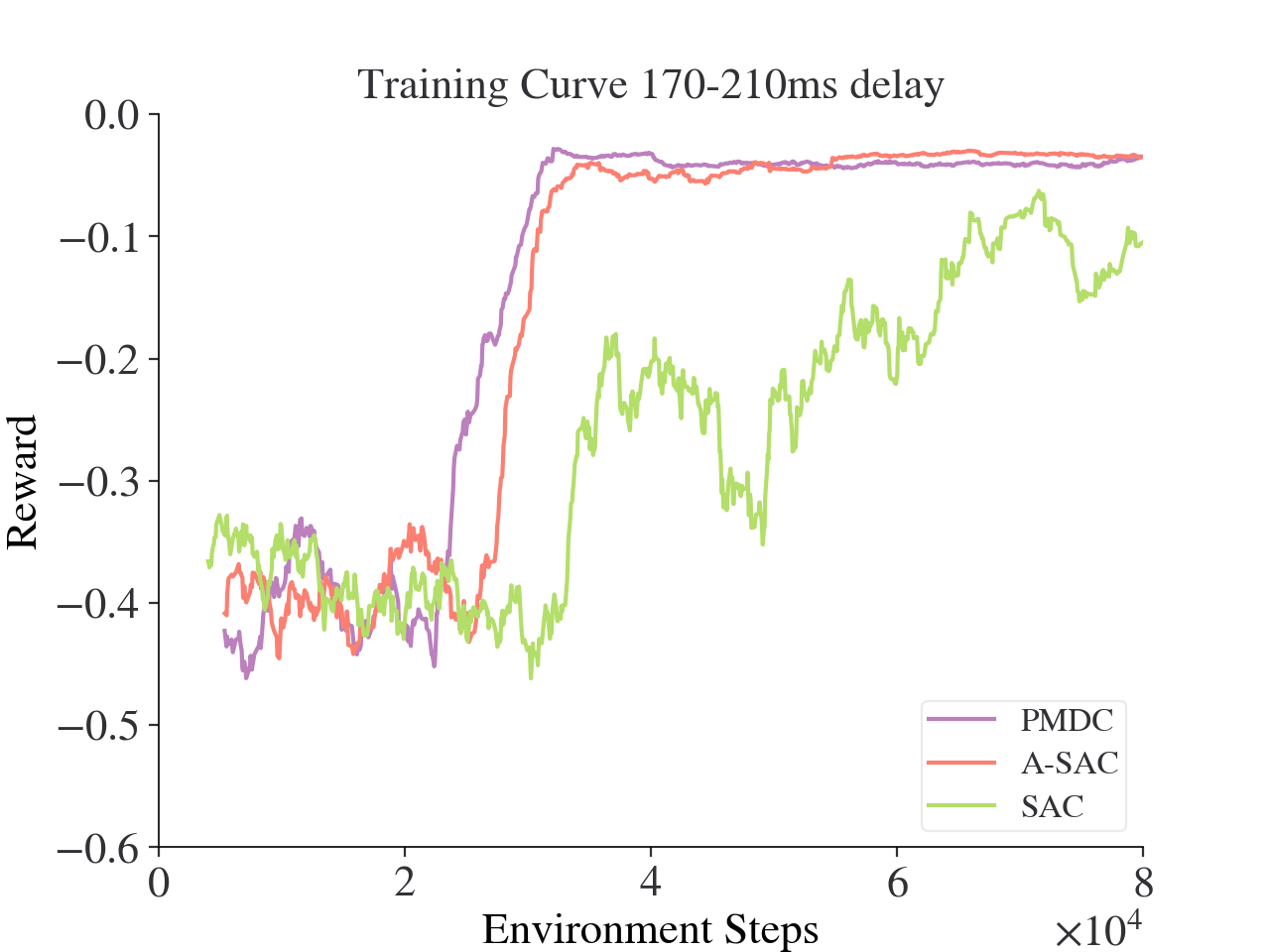} &
\includegraphics[trim=15mm 0mm 12mm 5mm, scale=0.39]{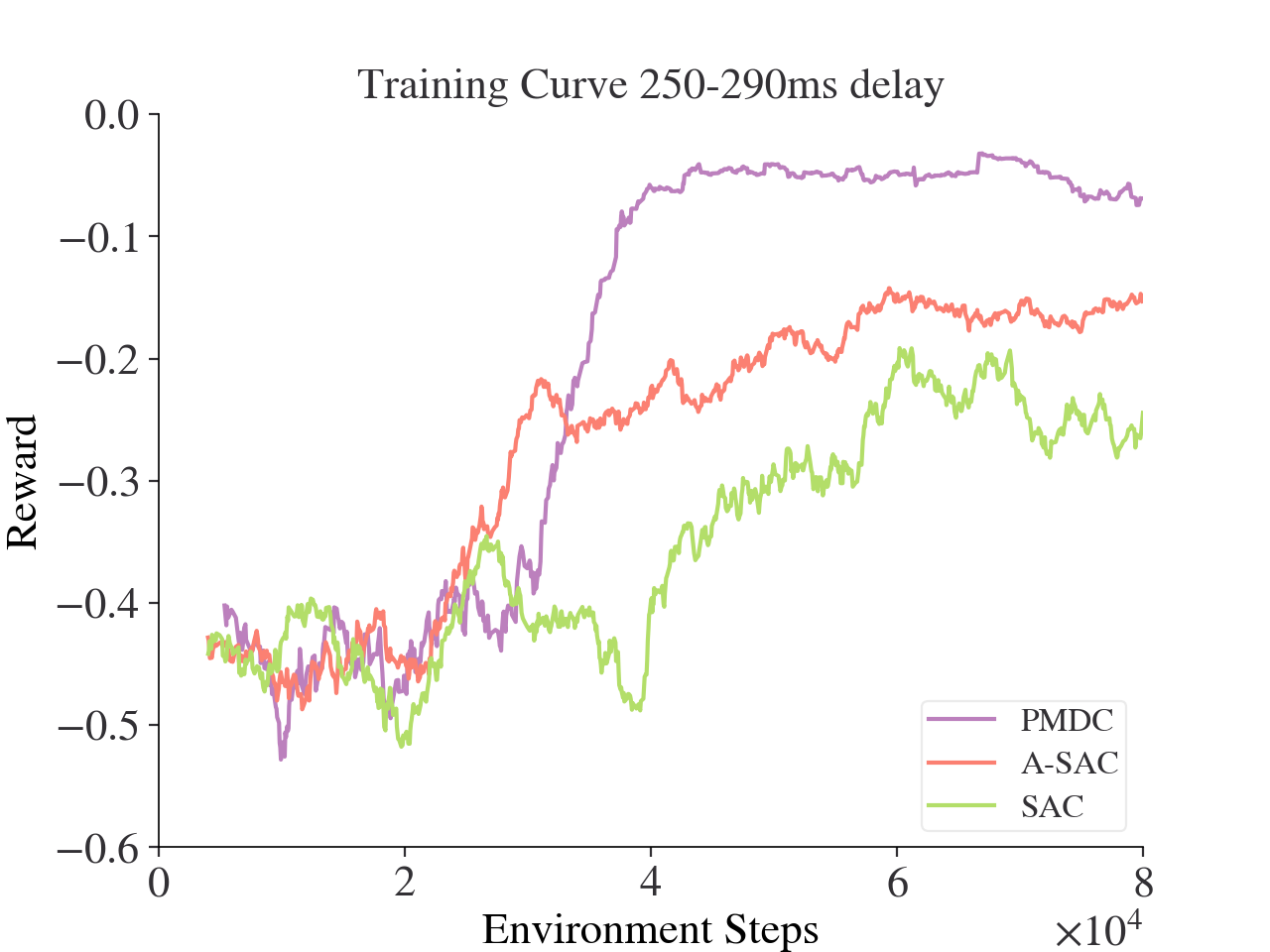} \\
(d) & (e) & (f)\\

\raisebox{-3.5mm}{\includegraphics[trim=15mm 7mm 5mm 0mm, clip,scale=0.45]{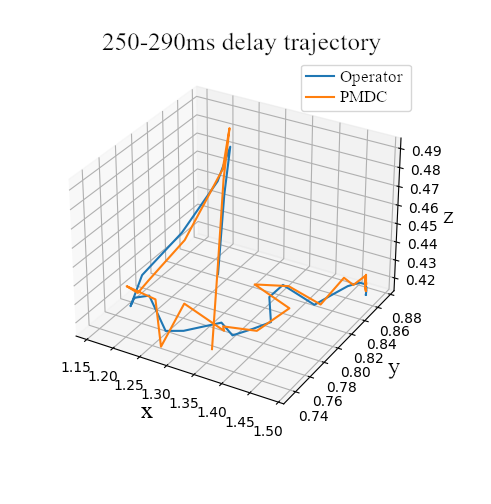}} & \includegraphics[trim=15mm 5mm 5mm 0mm, scale=0.45]{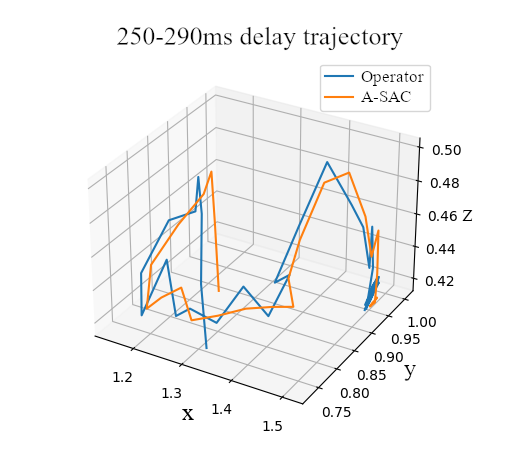} & \includegraphics[trim=15mm 5mm 5mm 0mm, scale=0.45]{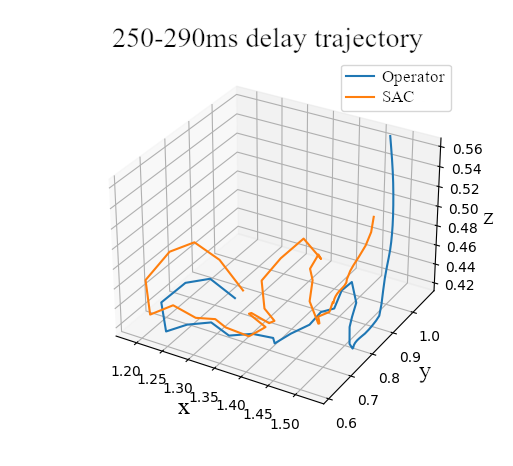} \\
(g) & (h) & (i)
\end{tabular}

\caption{
(a),(b),(c) Average training reward over Environment time-steps comparing state-prediction methods SBSP and ABSP under varying delays: (a) 90-130ms (b) 170-210ms (c) 250-290ms
(d),(e),(f) Average training reward over Environment time-steps for comparing RL algorithms PMDC, A-SAC and SAC under varying delays: (a) 90-130ms (b) 170-210ms (c) 250-290ms
(g),(h),(i) Best trajectory out of 10 testing episodes for each of the RL algorithms under 250-290ms delay: (a) PMDC, (b) A-SAC, and (c) SAC}

\label{fig:results}
\end{figure*}
\section{Experiments}
In this section, we compare PMDC with SAC and A-SAC. To conduct our experiments, we use the local-remote equivalent (See Fig. \ref{fig:lr-arch} of the \texttt{FetchPush-v1} MuJoCo environment \cite{1802.09464}. This environment  features a 7-DoF robot arm that must push an object to a specified target location. We chose this environment because it represents a challenging and realistic task for a local-remote system that involves frequent interactions with an object. The time between transitions is 10ms for this environment and therefore an 80ms delay would be represented by 8 delayed time steps and each episode is 50 time steps. Since the objective of pushing the block is replaced with minimising the distance between the two robotic arms, the results of this experiment generalises to other robotics tasks naturally.

Since action and observation delay are mathematically equivalent \cite{katsikopoulos2003markov}, we simplify our experiments by using constant action delays that are corrected by the predictive model, and stochastic observation delays that require state augmentation. The environment has constant action delays of 80ms, 160ms, and 240ms, and stochastic observation delays ranging from 10ms to 50ms. This delay range was selected to demonstrate the performance of our models under both short and long delays, ensuring a fair comparison. These experiments increase the constant delay, to demonstrate how the predictive model performs, since increasing the stochastic delay range only increases the augmented state space applied on top of PMDC.






\begin{table*}[ht]

\centering
    \begin{tabular}{c c c c}
    \hline
    \textbf{Method} & \textbf{90-130ms} & \textbf{170-210ms} & \textbf{250-290ms} \\
    \hline
    ABSP &  2122 & 2808 & 3504 \\
    SBSP (PMDC) & 1538 & 1560 & 1577 \\
    A-SAC & 1451 & 1461 & 1477 \\
    SAC & 1477 & 1450 & 1457
    \end{tabular}
\caption{Average time over 3 runs to train an agent for 80k time steps on an NVIDIA GeForce 3080}
\label{tab:newvsprior}

\end{table*}

\begin{table*}[ht]

\centering
\begin{tabular}{c c c c c}
\hline
\textbf{Algorithm} & \textbf{90-130ms} & \textbf{170-210ms} & \textbf{250-290ms} \\ [0.5ex] 
\hline

PMDC & \textbf{-0.030} $\pm$ \textbf{0.013} &
-0.038 $\pm$ 0.020 & 
\textbf{-0.043} $\pm$ \textbf{0.031}\\ 

A-SAC & -0.034 $\pm$ 0.018 &
\textbf{-0.034} $\pm$ \textbf{0.012} &
-0.15 $\pm$ 0.17\\

SAC & -0.053 $\pm$ 0.14 &
-0.24 $\pm$ 0.33 &
-0.25 $\pm$ 0.31\\

\end{tabular}
\caption{Mean and standard deviation over the final training episode}
\label{tab:final_performance}

\end{table*}

\subsection{Discussion}
In the context of the adaptive control task, the performance metrics of both SBSP and ABSP exhibit notable similarities. Nonetheless, as the delay intensifies, ABSP exhibits a marginal superiority over SBSP, attributed to its increased ability to manage changes in non-linearity. However, SBSP is able to gain a significant computational efficiency advantage by assuming changes in error are linear and correcting them in the recalibration phase. This local linearity assumption allows SBSP to reuse prior predictions with minimal degradation in performance Figure \ref{fig:results}(a),(b),(c).
The respective efficiencies of the algorithms are demonstrated by the duration required to complete 80k time steps, is shown in Table \ref{tab:newvsprior}. This efficiency emerges from the fact that each increment in delay corresponds to $1$ additional prediction for SBSP per episode, while ABSP necessitates the computation of $50$ more predictions per episode.

For smaller delays, both PMDC and A-SAC achieve similar performance, as the augmented state space remains manageable in terms of dimensionality. However, as the delay length increases, the differences between the approaches become more pronounced, with PMDC converging to an optimal policy much faster than SAC and A-SAC. The results shown in Fig. \ref{fig:results} show the faster convergence of PMDC over SAC and A-SAC in various delayed conditions for the task of synchronising two robotic manipulators whilst they complete the task of pushing a brick. Table \ref{tab:final_performance} shows that PMDC has comparable performance to A-SAC for small delays and superior final performance compared to A-SAC and SAC as delay increases.

The advantage of a predictive model for long delays is it can simulate transitions during the initial delay period, allowing for better use of time steps. Since it is always acting in a time delay corrected environment, it is able to see the results of $t$ interactions with the environment, whereas the typically delayed system will only experience $t-(\alpha+\omega)$ interactions, as the final actions it chooses will not be observed until after the end of the episode. This means the final $\alpha+\omega$ transitions are in the imagination space and not real environment transitions but predicted future outcomes.

The trajectories with the highest reward over 10 testing episodes under 250-290ms delays are shown in (g), (h), and (i). These trajectories demonstrate that the PMDC offers greater stability and can follow the path of the operator even under highly delayed conditions.

Table \ref{tab:newvsprior} demonstrates experimentally how SBSP's growth in computation time is negligible in comparison to ABSP.

\section{Conclusion}
This work introduced an adaptive PD controller using delay-corrected RL and evaluated its performance in the task of synchronising local-remote systems. This framework enables the training of adaptive PD controllers specialised in mapping local-remote system positions. We then demonstrated that through the use of a predictive dynamics model, we are able to increase the final performance of our controller further and accelerate convergence. This approach was evaluated against SAC and A-SAC, two approaches typically used to handle delayed control.

Future work can provide an analysis of the computation time against comparable model-based methods in various other delayed tasks. Another improvement may arise from experimentation with architectural differences such as RBF and probabilistic networks \cite{chua2018deep,chen2021delay} to further the performance. Similarly, another useful comparison would be against methods that attempt to condense the action history e.g. via a belief representation \cite{Liotet2021} or through a recurrent network.
Further work can also examine how to handle delays that are not a multiple of the time-step interval and experiment with real-world hardware or the effect of planning methods on performance and computation time.


\section*{Acknowledgments}
The first author thanks research funding support from UK Engineering and Physical Science Research Council (project ref: EP/T518050/1) and Veolia Nuclear Solutions
\printbibliography[heading=bibintoc]

\end{document}